\documentclass[10pt,twocolumn,letterpaper]{article}

\usepackage[pagenumbers]{cvpr} 
\usepackage{multirow}

\pdfoutput=1
\usepackage[dvipsnames]{xcolor}
\usepackage{float}
\usepackage{graphicx}
\usepackage{subcaption}
\usepackage{makecell}
\usepackage{booktabs}

\pdfoutput=1

\definecolor{cvprblue}{rgb}{0.21,0.49,0.74}
\usepackage[pagebackref,breaklinks,colorlinks,citecolor=cvprblue]{hyperref}
\usepackage{hyperref}
\hypersetup{
    colorlinks=true,
    linkcolor=red,
    urlcolor=magenta,
    citecolor=green
}

\def\paperID{341} 
\def\confName{CVPR}
\def\confYear{2024}

\title{SynFog: A Photo-realistic Synthetic Fog Dataset based on End-to-end Imaging Simulation for Advancing Real-World Defogging in Autonomous Driving}

\author{Yiming Xie\textsuperscript{1*}, Henglu Wei\textsuperscript{1*}, Zhenyi Liu\textsuperscript{2}, Xiaoyu Wang\textsuperscript{1}, Xiangyang Ji\textsuperscript{1}
\and
\textsuperscript{1}Tsinghua University, \textsuperscript{2}Stanford University
\and
{\tt\small \{xieym22@mails., weihenglu@mail., xiaoyu-w17@mails., xyji@\}tsinghua.edu.cn}, \\ \tt\small \{zhenyiliu@\}stanford.edu}

\begin{document}

\twocolumn[{
\renewcommand\twocolumn[1][]{#1}
\maketitle
\begin{center}
    \centering
    \includegraphics[width=0.98\textwidth]{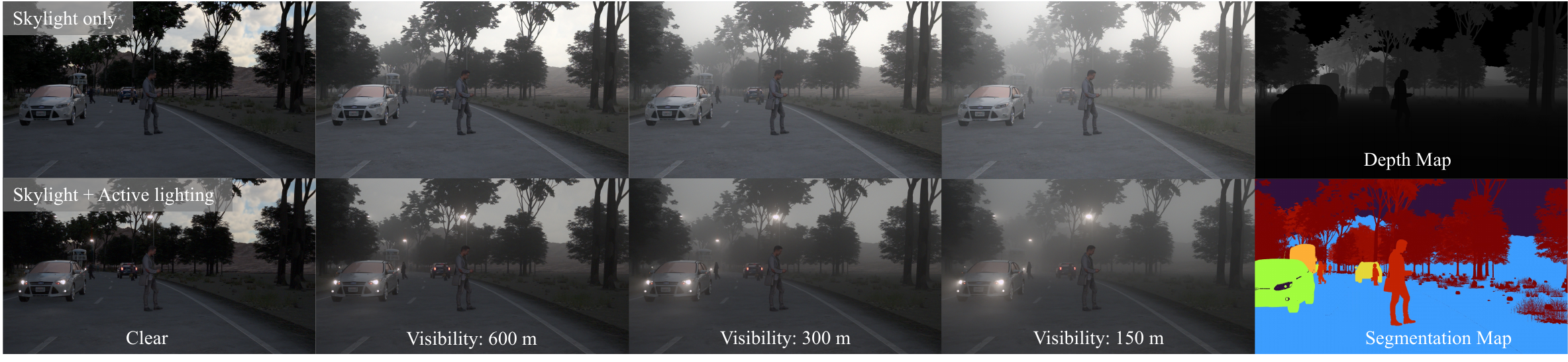}
    \captionof{figure}{The SynFog dataset comprises images under different lighting, with the first row captured in natural sky light and the second with extra light sources like street lamps and automotive lighting. This dataset covers three levels of fog density across each scene, corresponding visibility of 600 m, 300 m, 150 m. Additionally, pixel-accurate depth data and segmentation labels for each scene are also provided.}
    \label{fig:1.0_teaser}
    \vspace{-0.15cm}
\end{center}
}]

\begin{abstract}
\vspace{-0.3cm}
To advance research in learning-based defogging algorithms, various synthetic fog datasets have been developed. However, existing datasets created using the Atmospheric Scattering Model (ASM) or real-time rendering engines often struggle to produce photo-realistic foggy images that accurately mimic the actual imaging process. This limitation hinders the effective generalization of models from synthetic to real data. In this paper, we introduce an end-to-end simulation pipeline designed to generate photo-realistic foggy images. This pipeline comprehensively considers the entire physically-based foggy scene imaging process, closely aligning with real-world image capture methods. Based on this pipeline, we present a new synthetic fog dataset named SynFog, which features both sky light and active lighting conditions, as well as three levels of fog density. Experimental results demonstrate that models trained on SynFog exhibit superior performance in visual perception and detection accuracy compared to others when applied to real-world foggy images. 
\end{abstract}
\vspace{-0.65cm}   
\section{Introduction}
\label{sec:intro}
Fog, due to its light scattering and attenuation properties, poses a significant challenge for outdoor vision-based systems such as remote sensing, surveillance, autonomous driving, and fire rescue \cite{buch2011review, dai2010satellite, zhang2018vision}. It causes severe degradation in object appearance and contrast, resulting in a notable impact on people's daily lives. Consequently, there has been a growing research interest in image defogging task to develop robust outdoor vision systems.

\vspace{-0.05cm}

\par Recent learning-based defogging algorithms depend on sets of foggy images and corresponding clear images for model training. While several datasets have been collected from real-world scenarios \cite{sakaridis2018semantic, bijelic2020seeing}, they are typically limited to foggy images only, as it is almost impossible to obtain pixel-accurate clear images with consistent scene contents and environmental illumination. Studies in \cite{ancuti2018haze, ancuti2019dense, ancuti2020nh, narayanan2023multi} attempt to capture real-world foggy and clear image pairs using smoke machines. However, the scale of these datasets is inadequate for training of a high-performing model. Given the challenge of obtaining paired foggy and clear images in real world, various synthetic fog datasets have been proposed as an alternative \cite{li2018benchmarking, sakaridis2018semantic, zhang2017hazerd}. The most widely used method is based on the theory proposed by McCartney \cite{mccartney1976optics}, also known as the Atmospheric Scattering Model (ASM). However, this model fails to consider accurate global illumination and the actual imaging process that occurs in real-world captures as detailed in Sec. \ref{method:fog_model}. These limitations significantly impact the model's overall applicability, particularly in nighttime scenarios, and can lead to a disparity between synthetic and real-world foggy images. As a result, models trained on synthetic foggy images may exhibit limited robustness when applied to real-world ones.

\begin{figure}
        \centering
	 \includegraphics[width=0.5\textwidth]{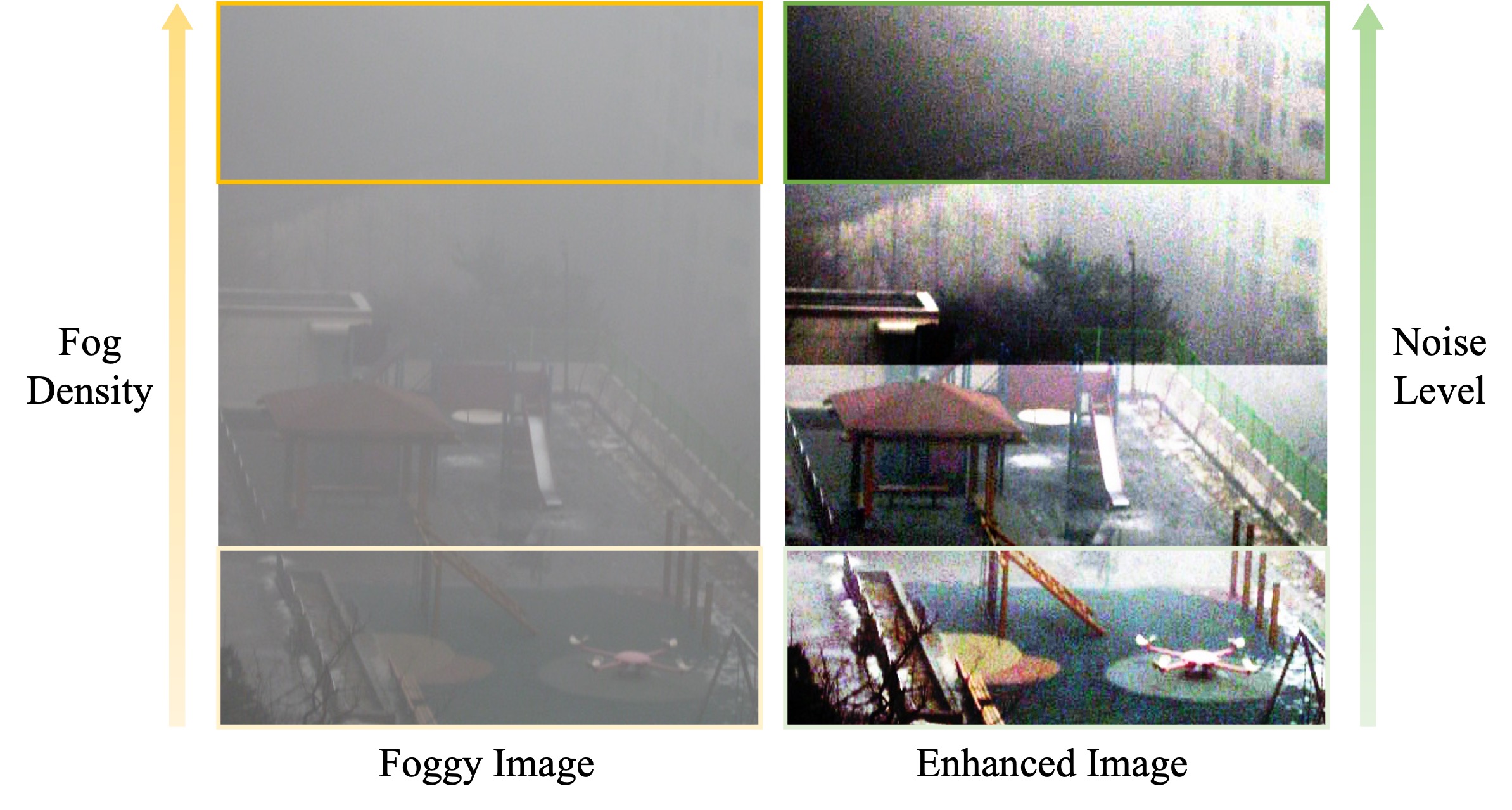}
	\caption{Relationship between noise level and fog density in real-world foggy images. Using regional contrast stretching for image enhancement, it can be observed that the noise level increases with the concentration of fog. \iffalse With the increase of fog density, regional contrast stretching enhancement method can no longer reveal the objects and details, which are buried amidst the rising noise levels.\fi}
	   \label{fig:1.1_noise_level}
    \vspace{-0.4cm}
\end{figure}

\par To address these issues, we propose an end-to-end foggy image simulation pipeline that encompasses the entire imaging process starting from 3D virtual scenes to final photorealistic foggy images. Our approach involves foggy scene rendering using volumetric path tracing and imaging through a realistic camera model. As shown in Fig. \ref{fig:1.1_volume}, fog is treated as a spatial volume, and the spectral scene radiance is rendered using volumetric path tracing \cite{pharr2023physically}. By defining the phase function of fog medium and bidirectional reflectance distribution function (BRDF) of the scene components, we are able to accurately describe the scattering process with multiple light sources under foggy conditions. The foggy scene radiance data is further processed through a physically-based camera model, which comprises optics, sensor and image processing, to faithfully replicate real camera devices. Based on the proposed pipeline, we develop a new extensive synthetic fog dataset called SynFog. This dataset contains 500 unique outdoor scenes, each featuring two kinds of lighting conditions and three levels of fog density. Extensive experiments have been conducted to qualitatively and quantitatively validate the authenticity of our simulation pipeline and the SynFog dataset.

\par Our main contributions are summarized as follows:
\begin{itemize}
    \item We propose a physically-based end-to-end foggy image simulation pipeline that incorporates accurate light transportation in scattering medium and physical characteristics of optics and sensor into the synthesized images.
    \item We develop a photo-realistic synthetic fog dataset, SynFog, which includes both sky light and active lighting conditions, as well as three levels of fog density.
    \item We demonstrate the authenticity and effectiveness of our simulation pipeline and the superior realism of SynFog dataset compared to other synthetic fog datasets.
\end{itemize}


\section{Related Work}
\label{sec:related_work}
\par Given the challenge of acquiring paired foggy and clear images in the real world for training learning-based defogging methods, researchers have proposed many synthetic fog datasets as an alternative. In this section, we will offer a thorough review and analysis of foggy image simulation techniques and their associated datasets. 
\par \textbf{Based on Atmospheric Scattering Model (ASM).} This method involves rendering fog onto clean images along with their depth information \cite{ancuti2016d, zhang2017hazerd, li2018benchmarking, sakaridis2018semantic, zhang2021simulation, hahner2019semantic, li2020realistic}. Depth information can be acquired through game engines, depth cameras or depth estimation methods. In an earlier study conducted by Tarel \textit{et al.} \cite{tarel2010improved}, FRIDA was constructed using a total of 90 synthetic images, with clear images and depth maps generated using \(SiVIC^{TM}\) software. Subsequently, in their later work \cite{tarel2012vision}, more synthetic images were added, leading to an upgraded version named FRIDA2, which includes 330 synthetic images. However, the scale of these datasets is insufficient for modern data-driven defogging methods, and the foggy images appear unrealistic. In more recent works \cite{zhang2021simulation, li2020realistic}, a similar approach has been employed to acquire both the clear images and their corresponding depth maps from game engines. In \cite{li2018benchmarking}, a general dehazing dataset called RESIDE was introduced. The indoor training set (ITS) of RESIDE was constructed based on existing indoor RGBD datasets NYUv2 \cite{silberman2012indoor} and Middleburry stereo \cite{scharstein2003high}. For the outdoor training set (OTS), 2071 real-world outdoor images were collected, and the corresponding depth information was estimated using a learning-based technique \cite{liu2015learning}. It is important to note that the accuracy of depth estimation based on learning methods is still unsatisfactory, as single-image-based depth estimation is inherently an ill-posed problem. Furthermore, Foggy Cityscapes was introduced in \cite{sakaridis2018semantic}, utilizing images from the Cityscapes dataset \cite{cordts2016cityscapes}. Despite efforts to enhance the original depth maps in Cityscapes, artifacts and discrepancies still persist, resulting in synthetic foggy images that appear unnatural. 

\par \textbf{Based on Real-time Rendering Engines.} Foggy images generated utilizing real-time rendering engines like Unity and Unreal have been discussed in works such as \cite{gaidon2016virtual, sun2022shift}. Among them, Virtual KITTI 2 \cite{gaidon2016virtual} stands out as a notable example. However, these foggy images often lack realism and exhibit limited diversity in terms of fog density. 

\par \textbf{Based on Physical Means.} Fog/haze machines are employed in real-world settings to generate artificial fog, as discussed in works such as \cite{ancuti2018haze, ancuti2019dense, ancuti2020nh, narayanan2023multi}. However, these datasets are generally limited in scale and may not offer sufficient supervision for learning-based defogging methods. Moreover, ensuring consistent content between the clear and foggy image captures can be challenging. 

\par There are also several real-world fog datasets available, such as Foggy Driving \cite{sakaridis2018semantic}, Foggy Zurich \cite{dai2020curriculum}, and Seeing Through Fog \cite{bijelic2020seeing}. However, due to the absence of corresponding clear reference images, these datasets can only be used for testing purposes rather than for training. In this paper, we propose a physically-based end-to-end foggy image simulation pipeline to improve the authenticity of synthetic foggy images and construct a photo-realistic fog dataset to facilitate future research on defogging algorithms.
\vspace{-0.15cm}
\section{End-to-end Foggy Image Simulation}
\label{sec:method}
\par
\begin{figure*}[t]
    \centering
	\includegraphics[width=\textwidth]{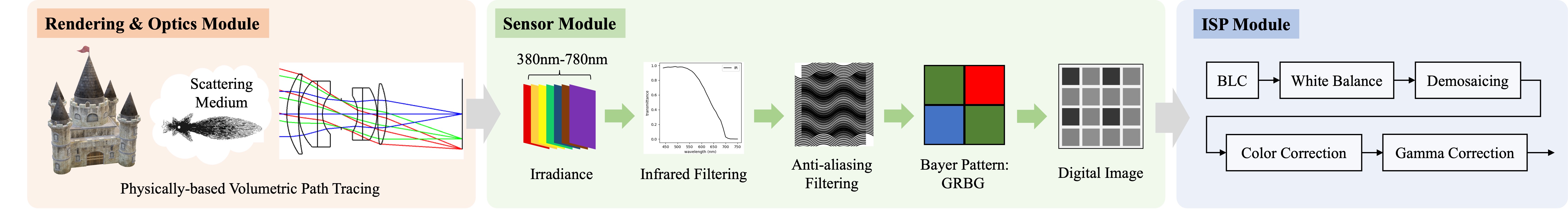}
	\caption{End-to-end foggy image simulation pipeline. The spectral radiance data is rendered using volumetric path tracing and passing through a realistic optics model before reaching the sensor plane. Subsequently, the irradiance is converted into a digital image through the sensor model. The raw image from the sensor is then processed by an ISP module to generate the final foggy image.}
	\label{fig:3.0_pipeline}
    \vspace{-0.3cm}
\end{figure*}

 We propose an end-to-end approach for generating photo-realistic foggy images that incorporates accurate light transportation in scattering medium and physical characteristics of optics and sensor into the synthesized images, as illustrated in Fig. \ref{fig:3.0_pipeline}. We first introduce a foggy scene imaging model in Sec. \ref{method:fog_model}. Based on this model, we establish a simulation pipeline consisting of two components: a) Foggy scene radiance is rendered using volumetric path tracing \cite{pharr2023physically}. b) The radiance data is processed through a physically-based camera model, which comprises optics, sensor and image processing to faithfully replicate real camera devices \cite{farrell2012digital}. These components are further detailed in Sec. \ref{method:pbrt} and Sec. \ref{method:camera_simulation}, respectively.

\begin{figure}[t]
    \centering
    \begin{subfigure}{0.4\textwidth}
        \centering
        \includegraphics[width=\textwidth]{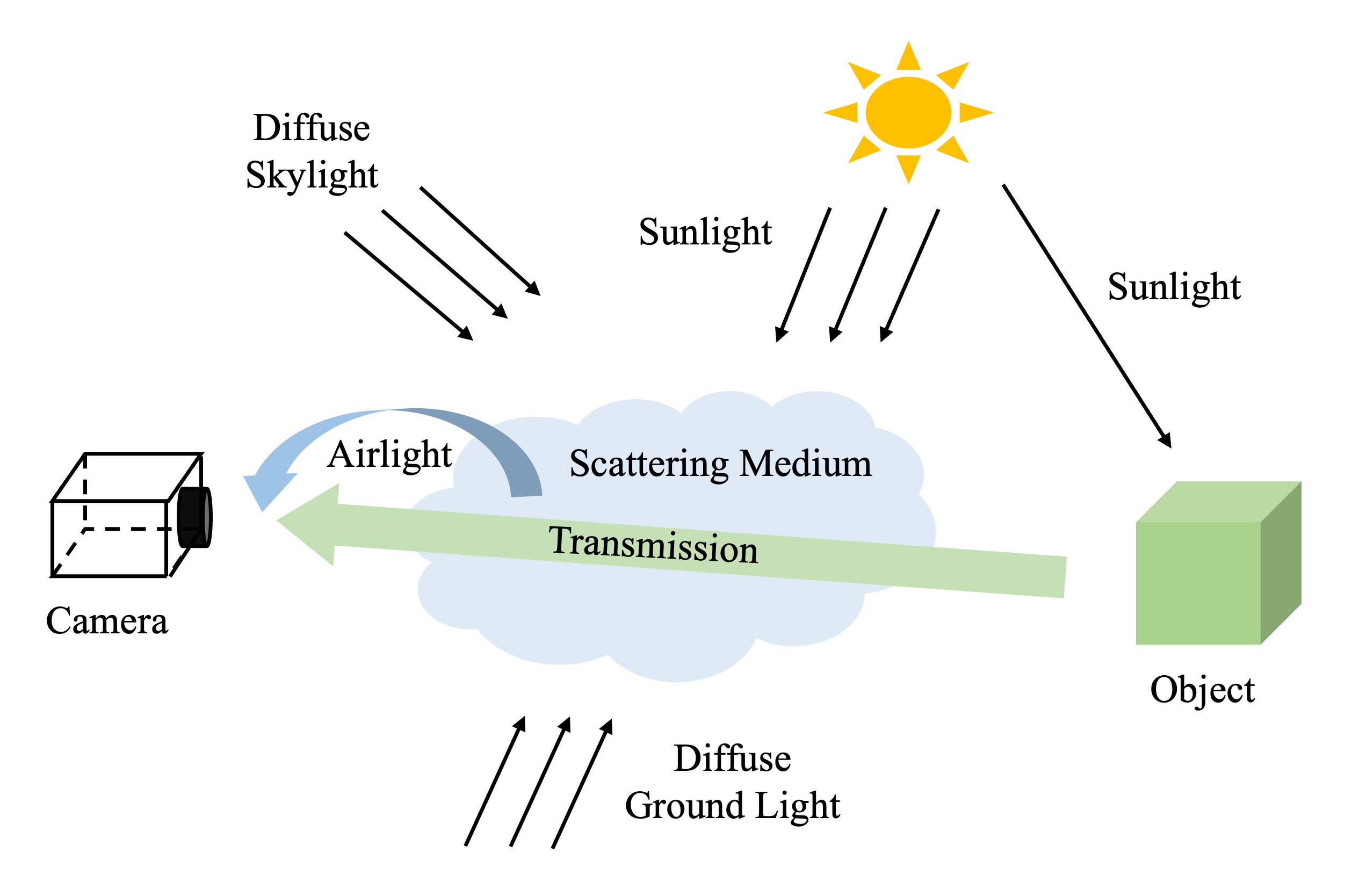}
        \caption{Atmospheric Scattering Model (ASM) \cite{narasimhan2002vision}.}
        \label{fig:1.1_ASM}
    \end{subfigure}
    
    \vspace{5pt}
    
    \begin{subfigure}{0.4\textwidth}
        \centering
        \includegraphics[width=\textwidth]{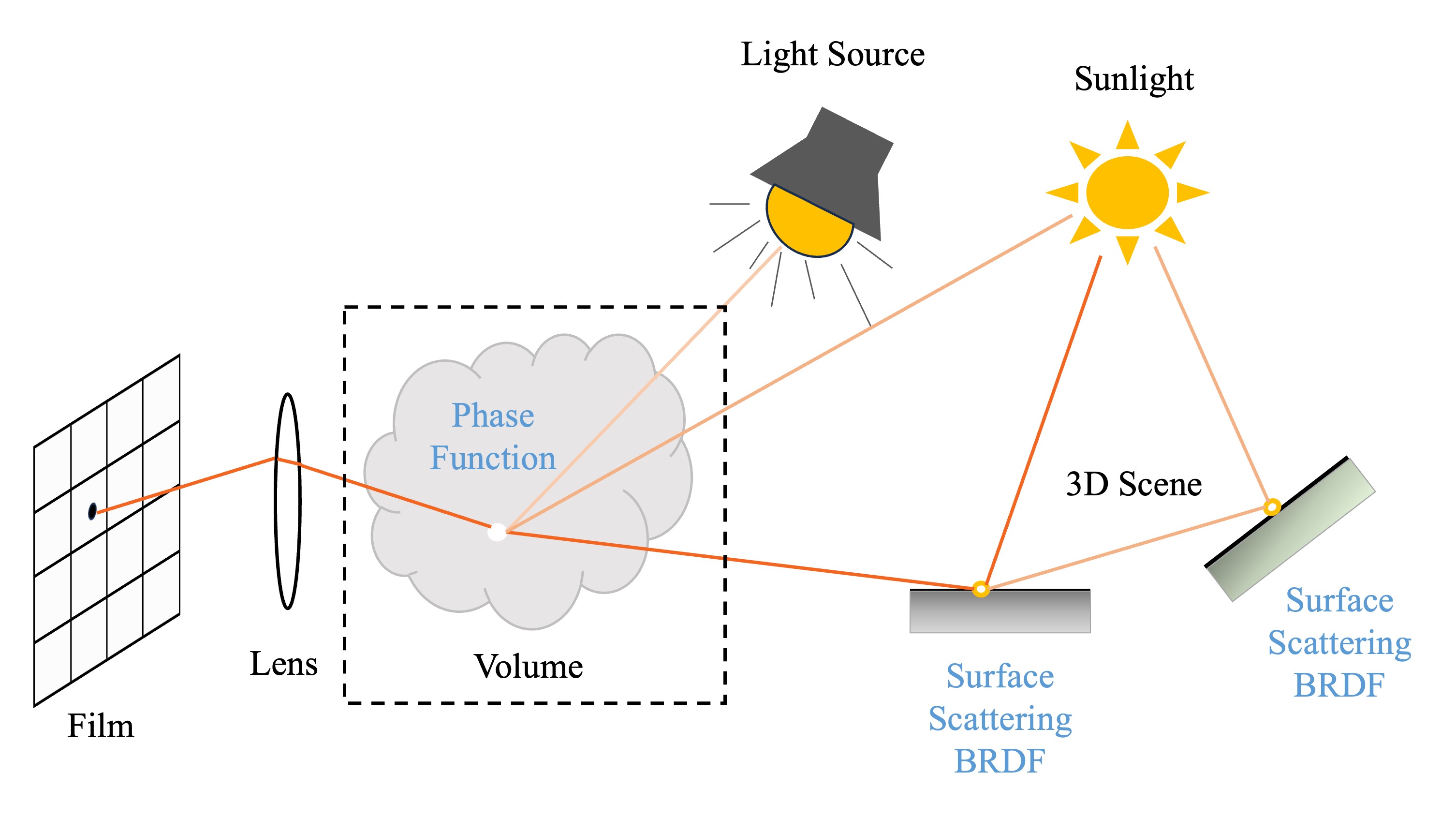}
        \caption{Foggy scene rendering based on volumetric path tracing.}
        \label{fig:1.1_volume}
    \end{subfigure}
    \caption{Foggy scene simulation methods.}
    \label{fig:fog model}
    \vspace{-0.5cm}
\end{figure}

\subsection{Foggy Scene Imaging Model}
\label{method:fog_model}

\par The Atmospheric Scattering Model (ASM) \cite{mccartney1976optics, narasimhan2002vision} has been widely used for the generation of foggy images \cite{sakaridis2018semantic, li2018benchmarking, hahner2019semantic, zhang2021simulation}:
\begin{equation}
E(d, \lambda)=\underbrace{E_{o}(\lambda) e^{-\beta(\lambda) d}}_{transmission} + \underbrace{L_{h}(\infty, \lambda)\left(1-e^{-\beta(\lambda) d}\right)}_{airlight},\label{JT_model}
\end{equation}
where \(E\) is the irradiance reaching the film under foggy conditions, \(E_{o}\) is the irradiance under clear conditions, \(\beta\) is the scattering coefficient characterizing the scatter ability of the medium, \(d\) is the distance between the object and the camera, \(L_{h}(\infty, \lambda)\) denotes the radiance of airlight at infinity and \(\lambda\) refers to wavelength. Fig. \ref{fig:1.1_ASM} provides a more intuitive description of this process. Foggy images can be generated with pre-defined \(\beta\) and \(L_{h}\) values by directly using the clear images as \(E_{0}\) and the corresponding depth maps as \(d\).

\par However, there are two issues with this scattering model for generating foggy images: (1) The definition of environmental illumination fails to consider the contribution of indirect illumination from the scene to the scattering medium, resulting in an unfaithful representation of the actual illumination. (2) The current formulation of airlight is insufficient for accurately modeling the lighting conditions that involve multiple point light sources, such as street lamps and automotive lights. This limitation restricts the model's applicability in situations like nighttime \cite{zhang2020nighttime, liu2022nighttime}. 

\par Furthermore, ASM only captures the irradiance reaching the sensor plane. Consequently, foggy images generated using ASM fail to accurately incorporate the realistic physical characteristics of optics and sensor in actual imaging process. In Fig. \ref{fig:1.1_noise_level}, we depict the relationship between noise level and fog density after the enhancement of regional contrast stretching \cite{treibitz2012resolution} on a real captured image \cite{choi2015referenceless}. It can be seen that the noise level increases with higher fog density, emphasizing the critical need for accurate noise modeling in foggy image simulation.

\par To enhance the consistency between synthetic images and real-world captures, we develop a comprehensive foggy scene imaging model that incorporates the entire imaging process, referred to as:
\begin{equation}
    I_{out} = f_{ISP}(f_{sensor}(f_{optics}(L_{scene})) + n),
    \label{foggy_imaging}
\end{equation}
where \(L_{scene}\) is the scene radiance obtained using volumetric path tracing \cite{pharr2023physically}, as described in Sec. \ref{method:pbrt}; \(\{f_{optics}, f_{sensor}, f_{ISP}\}\) denote the optics, sensor and ISP processes respectively, as described in Sec. \ref{method:camera_simulation}, and the noise term introduced by the sensor is denoted as \(n\).
\par By employing volumetric path tracing to obtain the radiance of foggy scene, we can precisely model the light scattering process under global illumination and active lighting. By incorporating the complete imaging process described in Eq. \eqref{foggy_imaging}, we are able to faithfully replicate the authentic camera characteristics in foggy image simulation.

\subsection{End-to-end Simulation Pipeline}
\label{method:pipeline}
\subsubsection{Foggy Scene Rendering}
\label{method:pbrt}
\par We utilize procedural modeling techniques, similar to the approach employed in \cite{tsirikoglou2017procedural,liu2019system}, to create large-scale and diverse driving scenes. This involves defining adjustable parameters and using a set of predefined rules to generate explicit scene definitions. Our model library incorporates a diverse range of pedestrians, bicyclists and cars, each with well-defined material and texture properties. Furthermore, we carefully select 16 sky maps with cloudy conditions to ensure realistic lighting for foggy scenes, following the established guidelines from previous works \cite{ancuti2018haze, ancuti2019dense, ancuti2020nh}.

\par In our approach, fog is treated as a spatial volume, and the spectral scene radiance is rendered using volumetric path tracing \cite{pharr2023physically}. To accurately model the scattering process of the fog medium, we employ Henyey and Greenstein phase function \cite{henyey1941diffuse, pharr2023physically} to model its particle scattering characteristics, denoted as:
\begin{equation}
p_{\mathrm{HG}}(\cos \theta)=\frac{1}{4 \pi} \frac{1-g^{2}}{\left(1+g^{2}+2 g(\cos \theta)\right)^{3 / 2}}. 
\end{equation}
By empirically setting the asymmetry parameter \(g=0.87\), we can accurately simulate the probability distributions for scattering in a specific direction at a given point within the fog medium, as shown in Fig. \ref{fig:1.1_volume}.

\par The rendering equation consists of two parts that account for reduction of radiance caused by out-scattering and increase of radiance due to in-scattering at point \(\mathrm{p}\) in direction \(\omega\), represented by Eq. \eqref{attenuation} and Eq. \eqref{airlight}, respectively \cite{pharr2023physically}.
\begin{equation}
     L_{\mathrm{i}}(\mathrm{p}^{\prime}, -\omega) = \mathrm{e}^{-\int_{0}^{d} \sigma_{\mathrm{s}}(\mathrm{p}+t \omega, \omega) \mathrm{d} t} L_{\mathrm{o}}(\mathrm{p}, \omega). \label{attenuation}
\end{equation}
\begin{equation}
    L_{\mathrm{s}}(\mathrm{p}, \omega) = L_{\mathrm{e}}(\mathrm{p}, \omega) + \sigma_{\mathrm{s}}(\mathrm{p}, \omega) \int_{\mathcal{S}^{2}} p\left(\mathrm{p}, \omega_{\mathrm{i}}, \omega\right) L_{\mathrm{i}}\left(\mathrm{p}, \omega_{\mathrm{i}}\right) \mathrm{d} \omega_{\mathrm{i}}. \label{airlight}
\end{equation}
\(\sigma_s\) represents the scattering probability per unit distance, \{\(L_{i}\), \(L_{o}\), \(L_{e}\), \(L_{s}\)\} are the \{incident, exitant, emission, total added\} radiance. In our implementation, we employ the Monte Carlo algorithm to solve the above integration. 

\par The key distinction between our physically-based foggy scene rendering technique and ASM lies in the modeling of airlight, as represented by Eq. \eqref{airlight}. Our approach offers a more accurate depiction of light scattering process through the utilization of specific phase function, and enables incorporation of global illumination and multiple light sources within the scene, producing authentic scene radiance to the camera model described in the next section.

\begin{figure}[t]
    \centering
	\includegraphics[width=0.48\textwidth]{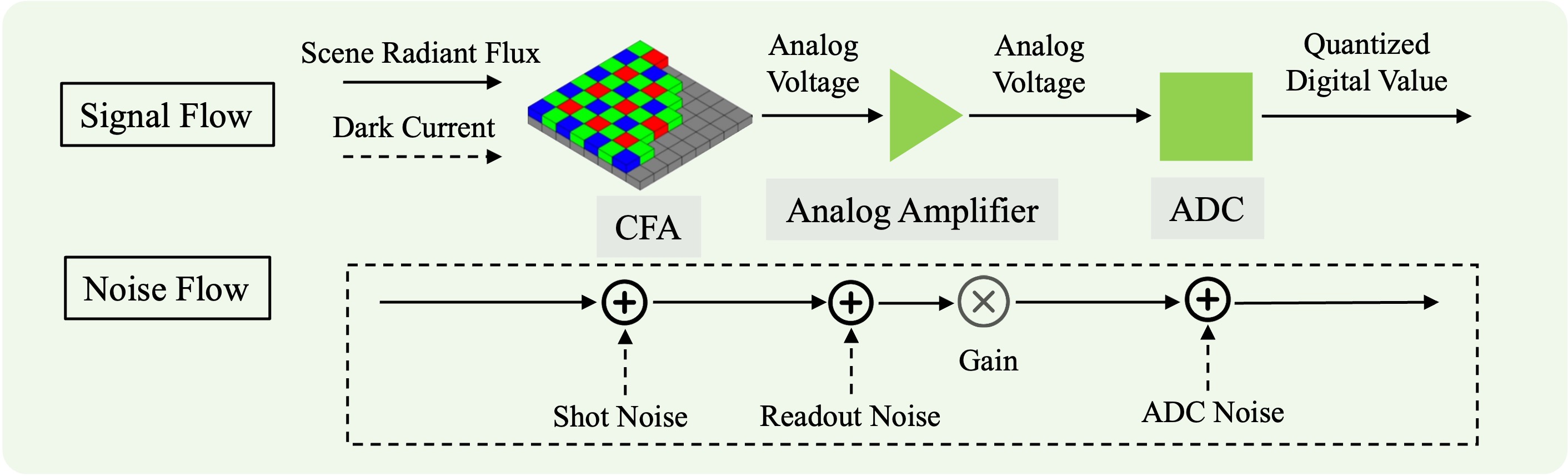}
	\caption{Noise flow and sensor simulation.}
	\label{fig:3.2_noise_flow}
        \vspace{-0.3cm}
\end{figure}

\vspace{-0.1cm}
\subsubsection{Imaging through Realistic Camera Simulation}
\label{method:camera_simulation}
\par After obtaining the scene radiance, we employ ISETCam \cite{farrell2003simulation, farrell2012digital, farrell2015image,liu2019soft,lyu2022validation} to simulate the realistic digital camera system that encompasses optics, sensor and ISP components. Specifically, we use a diffraction-limited optics model \cite{chen2003digital} with consideration of off-axis irradiance attenuation based on the cosine-fourth law \cite{klein2013optik} to convert scene radiance \(L_{o}\) to optical image irradiance \(E_{i}\), denoted as:
\begin{equation}
    E_{i}=\pi \cdot L_{o} \frac{1}{1+4(f / \#(1-m))^{2}}(\cos \phi)^{4},
\end{equation}
where \(f / \#\) is the f-number, \(m\) represents the magnification defined as the negative ratio of image distance and object distance. In this optics model, blur mainly depends on the f-number value but little else.

\par We model the sensor process including photovoltaic conversion, analog amplifier, and analog-to-digital conversion as described in Fig. \ref{fig:3.0_pipeline} and simulate the real noise characteristics depicted in Fig. \ref{fig:3.2_noise_flow}. Among the various noise sources, shot noise arising from the photovoltaic conversion emerges as the dominant type of noise in foggy images which follows the Poisson distribution. To ensure the credibility of our sensor simulation, we incorporate a set of calibrated sensor parameters from the Mi10Pro cellphone, encompassing the CFA and infrared filter data.

\par Finally, we process the RAW image output from the sensor with a simple ISP module, including black level compensation, white balance, demosaicing, color correction and gamma correction, to get the final synthetic image that mimics the real camera characteristics.
\section{SynFog Dataset}
\label{method:synfog_dataset}

\begin{table}[t]
    \centering
    \caption{Comparisons between SynFog and existing synthetic fog datasets in the driving field.}
    \label{table:comparison_between_datasets}
    \begin{tabular}{ccc}
        \bottomrule
        \multirow{2}{*}{Datasets} & Global & Optics and Sensor \\
        &  Illumination &  Characteristics \\
        \hline
        Foggy Cityscapes & \textcolor{red}{\(\times\)} & \textcolor{red}{\(\times\)} \\
        Virtual KITTI 2 & \textcolor{red}{\(\times\)} & \textcolor{red}{\(\times\)} \\
       SynFog (Ours) & \textcolor{green}{\(\checkmark\)} & \textcolor{green}{\(\checkmark\)} \\
        \toprule
    \end{tabular}
    \vspace{-0.3cm}
\end{table}

\par By utilizing the proposed pipeline, we develop a new extensive synthetic fog dataset called SynFog. Example images are shown in Fig. \ref{fig:1.0_teaser}. We first generate 500 large-scale, diverse driving scenes using procedural modeling techniques. For each scene, we generate two sets of images under different lighting conditions: one set with sky light only and the other set with both sky light and active light sources, such as street lamps and automotive lighting. Within each group, we create a set of four images: one clear image and three foggy images with varying fog levels defined by the scattering coefficient \(\sigma_{\mathrm{s}}\). According to McCartney's work \cite{mccartney1976optics}, the scattering coefficient remains independent of the wavelength for fog \cite{narasimhan2002vision, farrow1970influence}. Therefore, we use a consistent scattering coefficient for wavelengths within the visible range. To ensure realistic simulation of fog, we employ the meteorological optical range (MOR) \cite{kim2018comparison} or visibility as a reference to select appropriate scattering coefficients, described as:
\begin{equation}
    MOR=2.996/\sigma_{s}.
\end{equation}
Based on this equation, we select scattering coefficient values of 0.005, 0.01, 0.02, which correspond to visibility range of 600 m, 300 m and 150 m, respectively. According to National Standard of China \cite{grade_of_fog_forecast}, these visibility ranges are classified as heavy fog, thick fog, and dense fog, thereby the simulated foggy images are diversified enough to reflect real-world complicated foggy conditions. 

\par We compare SynFog with existing fog datasets in the field of autonomous driving, including the extensively used Foggy Cityscapes \cite{sakaridis2018semantic} and Virtual KITTI 2 \cite{gaidon2016virtual}. The comparison is presented in Tab. \ref{table:comparison_between_datasets}. In terms of foggy image generation, methods that rely on ASM \cite{narasimhan2002vision} require accurate depth maps to calculate the correct transmission maps. However, obtaining accurate depth maps for outdoor images can be challenging, leading to artifacts such as holes and discontinuities in synthetic foggy images, similar to what is observed in Foggy Cityscapes \cite{sakaridis2018semantic}. On the other hand, foggy images generated using Game Engines like Unity or Unreal \cite{gaidon2016virtual} typically employs rasterization techniques in the rendering process. While effective for creating scenes, rasterization techniques may not accurately model light transportation as precisely as path tracing, which is employed in SynFog. Moreover, SynFog takes the realistic imaging process into account, resulting in highly authentic foggy images with realistic physical characteristics. The superior realism of SynFog is further supported by the experimental results.

\par Furthermore, our data simulation pipeline offers exceptional flexibility and control, allowing us to generate not only the final RGB images but also other valuable data modalities, including RAW images. This capability opens up avenues for research on defogging algorithms in the RAW domain, which has not been explored due to the lack of paired foggy RAW images and clear RGB images necessary for training deep defogging networks.
\section{Experiments}
\label{sec:experiment}
\subsection{Fog Chamber Validation}
\label{exp:fog_chamber}
\par We establish an indoor fog chamber to validate the fidelity of our fog simulation pipeline. The chamber is constructed as a capped glass cube with dimensions of \(1.5m \times 0.5m \times 0.5m\). To reproduce the dark night environment that has directional light sources in the fog chamber, we position 4 light bars on each side of the front face of the cuboid fog chamber and enclose it with black blackout cloth from the inside. Fog inside the chamber is generated using a 1.7MHz ceramic atomizer that emulates water mist. To ensure uniformity of fog concentration within the chamber, we incorporate dual fans to promote even distribution. A visibility measurement module is designed to quantify the fog density. This module is comprised of a PM-160 wireless power meter from Thorlabs, an 852 narrowband laser, an aperture and a narrowband filter adapted to the laser band. The aperture serves to filter out stray light and the narrowband filter plays the role in blocking interference caused by non-laser light sources in the scene. The principle of visibility (scattering coefficient \(\sigma_{s}\)) measurement is expressed as \cite{nebuloni2005empirical}:  
\begin{equation}
    \sigma_{s}=\frac{\ln \left(\frac{P_{0}}{P_{u}}\right)}{\mathrm{u}},
\end{equation}
where \(P_{0}\) is the laser power measured without fog and \(P_{u}\) is the measured laser power after the propagation of \(u\) distance under foggy conditions.

\par We then use Blender to create a virtual fog chamber that can reproduce the real-world counterpart. Our process involves calibrating the light conditions, object information and camera properties prior to capture, resulting in a one-to-one reproduction of the real fog chamber. We select the standard Macbeth color checker as our test object and use Mi10Pro cellphone to capture images. We evaluate the matching degree between the simulation and reality under clear conditions by conducting a color analysis on the checker, which is available in the supplementary material.

\par After ensuring the consistency of initial capture conditions, we proceed to capture images under foggy conditions. We first record the laser power under clear condition as \(P_{0}\), and then proceed to generate fog until the object is no longer visible. During the fog dispersion process, we capture images at regular intervals of 1 second and record the laser energy \(P_{u}\) to obtain real-time fog density measurements. We utilize the measured fog density parameter \(\sigma_{s}\) to simulate corresponding foggy counterparts based on the proposed foggy image simulation pipeline. 

\par
\begin{figure}
    \centering
    \begin{subfigure}{0.48\textwidth}
        \centering
        \includegraphics[width=\textwidth]{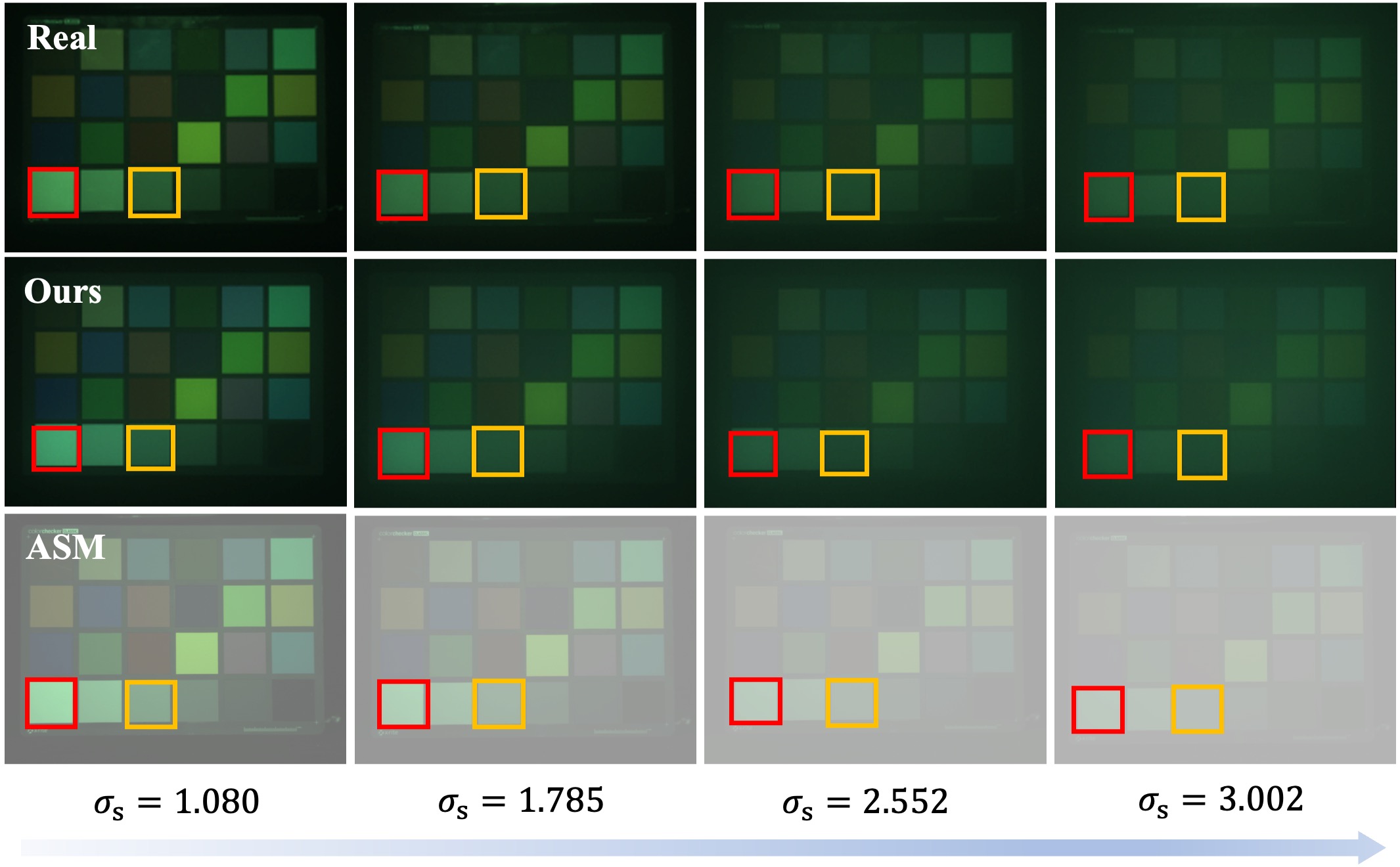}
        \caption{}
        \label{fig:4.1_foggy_match}
    \end{subfigure}
    
    \vspace{5pt}
    
    \begin{subfigure}{0.48\textwidth}
        \centering
        \includegraphics[width=\textwidth]{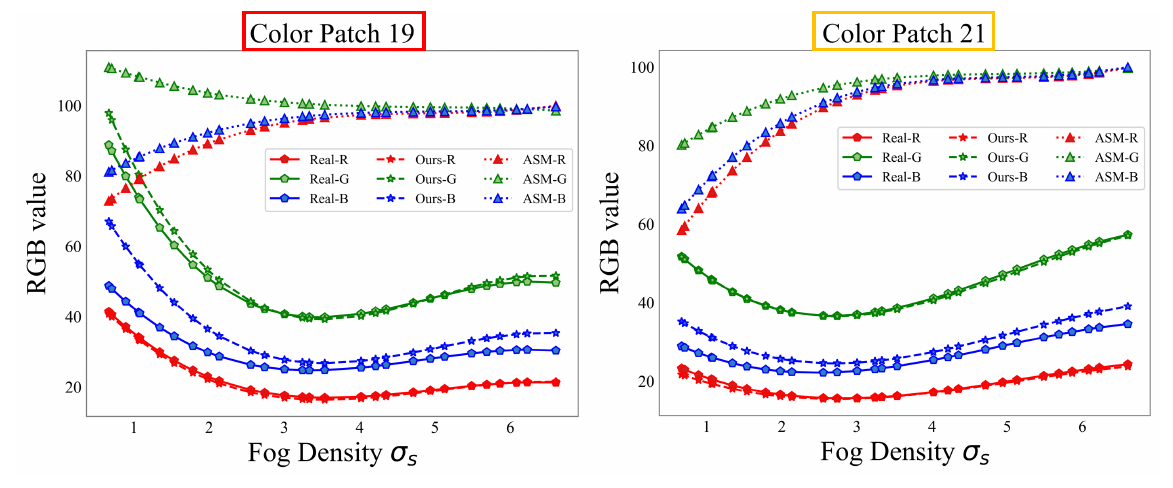}
        \caption{}
        \label{fig:4.1_trend_reality}
    \end{subfigure}
    \caption{Color analysis under foggy conditions. (a) Visual comparison between real captures and simulated images under different fog densities in linear RGB space. We increase the brightness of the images (\(\times\)2) for visual purpose. (b) Analyzing the trend of color variation across different fog densities.}
    \label{fig:4.1_foggy_match_trend_reality}
    \vspace{-0.5cm}
\end{figure}

 In Fig. \ref{fig:4.1_foggy_match}, we showcase both the real-captured foggy images and simulated ones generated using our pipeline and the ASM \cite{mccartney1976optics}. It can be seen that images simulated through our pipeline exhibit a closer resemblance to real-captured foggy images. We also conduct a color analysis on 24 color patches under different fog densities as depicted in Fig. \ref{fig:4.1_trend_reality}. The results demonstrate that, as fog density increases, the trend of each channel in simulated images generated by our pipeline consistently follows that of the real captures. In contrast, the simulated images based on ASM do not exhibit this trend. These findings provide a quantitative validation of the realism of our foggy image simulation method.

\begin{table*}[h]
    \centering
    \setlength{\abovecaptionskip}{0.1cm}
    \caption{Defogging evaluations with AECRNet on real-world fog datasets.}
    \label{table:dehaze}
    \begin{tabular}{c|ccc|c|c|c}
        \bottomrule
         \multirow{2}{*}{Training Set} & \multicolumn{3}{c|}{O-Haze \cite{ancuti2018haze}} & \multicolumn{1}{c|}{Foggy Zurich \cite{dai2020curriculum}} & \multicolumn{1}{c|}{Foggy Driving \cite{sakaridis2018semantic}} & \multicolumn{1}{c}{BeDDE \cite{zhao2020dehazing}}\\
         \cline{2-7}
         & PSNR \(\uparrow\) & SSIM \(\uparrow\) & DHQI \cite{min2018objective} \(\uparrow\) & DHQI \cite{min2018objective} \(\uparrow\) & DHQI \cite{min2018objective} \(\uparrow\) & DHQI \cite{min2018objective} \(\uparrow\)\\
        \hline
        Foggy Cityscapes  & 14.46 & 0.5737 & 43.40 & 52.06 & 51.55 & 36.07\\
        Virtual KITTI & 13.90 &  0.5315 & 42.80 & 50.94 & 47.46 & 33.42\\
        \textbf{SynFog}  & \textbf{15.43} & \textbf{0.6116} & \textbf{44.46} & \textbf{54.16} & \textbf{52.07} & \textbf{43.28}\\
        \toprule
    \end{tabular}
    \vspace{-0.2cm}
\end{table*}

\begin{figure*}[t]
        \centering
	\includegraphics[width=0.95\textwidth]{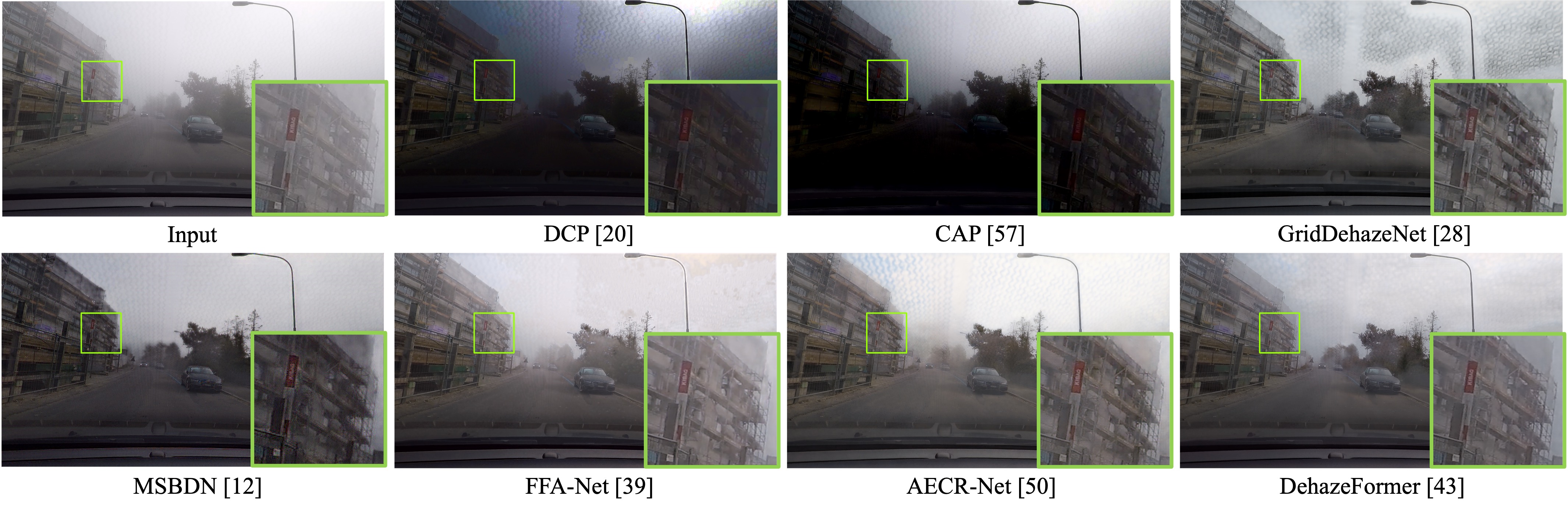}
    \vspace{-0.2cm}
	\caption{Transferability across real-world fog datasets. Test results of models trained on SynFog.}
	\label{fig:4.3_RealWorldTest}
    \vspace{-0.45cm}
\end{figure*}

\subsection{Transferability across the Real-to-Virtual Gap}
\label{exp:transferability}
\par We evaluate 7 representative state-of-the-art defogging algorithms on SynFog dataset: DCP \cite{he2010single}, CAP \cite{zhu2015fast}, GridDehazeNet \cite{liu2019griddehazenet}, MSBDN \cite{dong2020multi}, FFA-Net \cite{qin2020ffa}, AECR-Net \cite{wu2021contrastive}, and DehazeFormer \cite{song2023vision}. We directly apply the models trained on SynFog to real foggy images sourced from Foggy Zurich \cite{dai2020curriculum} and Seeing Through Fog \cite{bijelic2020seeing} datasets, without undergoing any model fine-tuning or domain adaptation. During the training process, we adhere to the official settings of these methods, and all of them demonstrate convergence by the end of the training. The results are shown in Fig. \ref{fig:4.3_RealWorldTest}. Despite DCP \cite{he2010single} and CAP \cite{zhu2015fast} producing excessively dark outputs, models trained on SynFog can produce natural defogged images with minimal artifacts.

\par To underscore the superior realism of SynFog compared to existing synthetic fog datasets, we further enhance our evaluation by training the top-performing models (AECR-Net \cite{wu2021contrastive}, DehazeFormer \cite{song2023vision}) on two prominent synthetic fog datasets relevant to autonomous driving (Foggy Cityscapes \cite{sakaridis2018semantic}, Virtual KITTI \cite{gaidon2016virtual}). Subsequently, these models are evaluated on real fog datasets for a comparative analysis. Additionally, we utilize ASM \cite{narasimhan2002vision} to create a SynFog-\(\beta\) dataset with clear images from SynFog and test it using DehazeFormer \cite{song2023vision}.

\begin{figure}[t]
    \centering
    \setlength{\abovecaptionskip}{0.2cm}
	\includegraphics[width=0.48\textwidth]{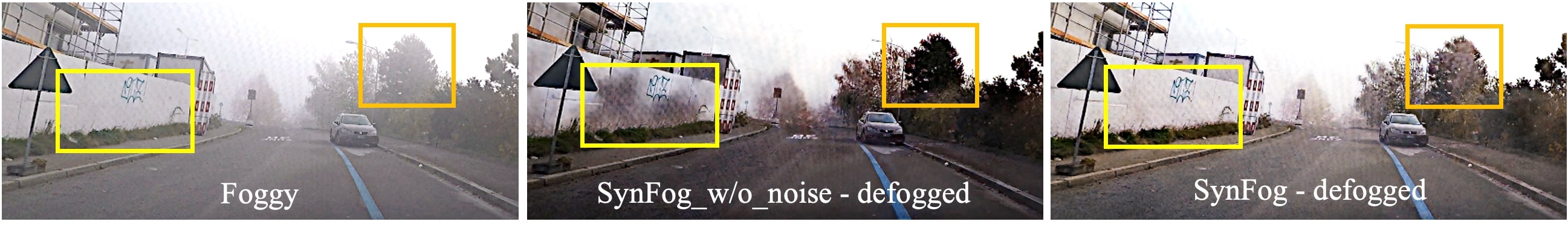}
        \vspace{-0.4cm}
        \caption{Ablation of sensor noise simulation.}
    \label{fig:ablation}
    \vspace{-0.2cm}
\end{figure}

\begin{table}[t]
    \centering
    \small
    \setlength{\abovecaptionskip}{0.1cm}
    \caption{Detection evaluations on real-world fog datasets. ``FZ" refers to Foggy Zurich. ``STF" represents Seeing Through Fog.}
    \label{table:detection}
    \begin{tabular}{c|c|c|c}
        \bottomrule
        \multirow{2}{*}{Method} & \multirow{2}{*}{Training Set} & \multicolumn{1}{c|}{FZ \cite{dai2020curriculum}} &  \multicolumn{1}{c}{STF \cite{bijelic2020seeing}}\\
         \cline{3-4}
        & & mAP (\%) & mAP (\%)\\
        \hline
        \multirow{3}{*}{AECRNet} & Foggy Cityscapes & 69.7 & 54.8\\
        & Virtual KITTI & 68.9 & 53.3\\
        & \textbf{SynFog} & \textbf{71.5} & \textbf{55.5}\\
        \hline
        \multirow{4}{*}{DehazeFormer} & Foggy Cityscapes & 67.9 & 54.9\\
        & Virtual KITTI & 68.5 & 53.1\\
        & SynFog-\(\beta\) & 59.7 & \textbf{55.3}\\
        & \textbf{SynFog} & \textbf{69.7} & \textbf{55.3}\\
        \toprule
    \end{tabular}
    \vspace{-0.6cm}
\end{table}

\par
\begin{figure*}[t]
        \centering
	\includegraphics[width=0.98\textwidth]{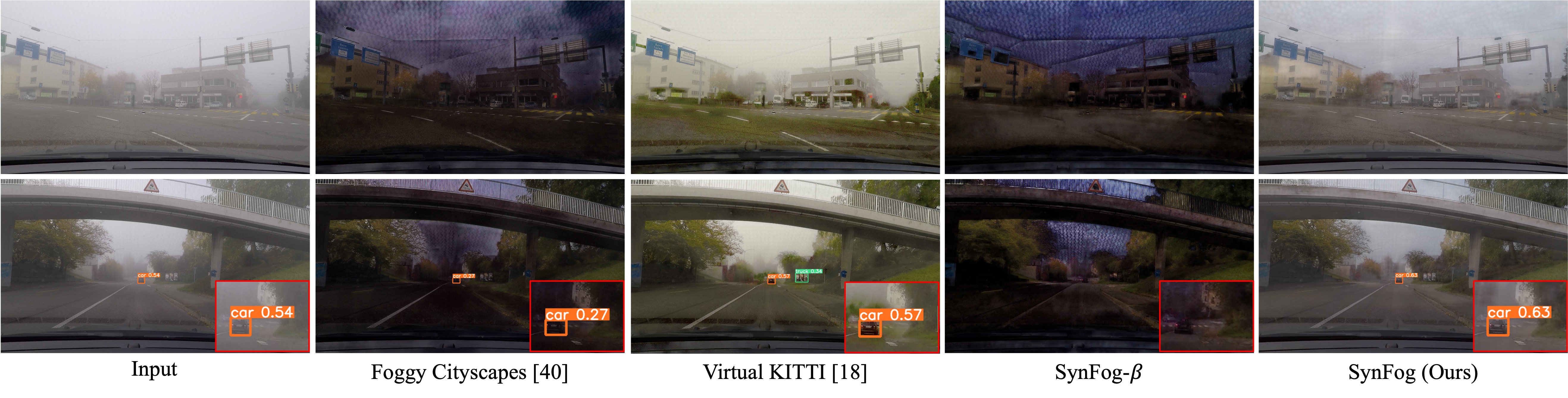}
	\caption{Visual and detection comparisons on real-world foggy images using DehazeFormer models trained on SynFog, SynFog-\(\beta\), Foggy Cityscapes \cite{sakaridis2018semantic} and Virtual KITTI \cite{gaidon2016virtual}.}
	\label{fig:4.4_ablation}
    \vspace{-0.1cm}
\end{figure*}

 \textbf{Comparison with existing synthetic fog datasets.} To mitigate any potential impact arising from model training issues, we initially conduct tests for each model using the corresponding test set of the training data. Test results are available in the supplementary material. Both AECR-Net \cite{wu2021contrastive} and DehazeFormer \cite{song2023vision} exhibit high performance, confirming the correctness of the model training process. Subsequently, the trained models are directly applied to real-world test foggy images. Results are shown in Fig. \ref{fig:4.4_ablation} and Tab. \ref{table:dehaze}. Owing to the high authenticity of the SynFog dataset, models trained on it can effectively generalize to real foggy images and produce naturally colored defogged images. Conversely, models trained on the other two datasets display poorer generalization to real data, resulting in artifacts and color distortion in defogged images. We also evaluate the defogged images using a pre-trained YOLOv8 \cite{yolov8} model for detection, showcasing how SynFog enhances machine vision capabilities (See Tab. \ref{table:detection}). More results can be found in the supplementary material.
\par It is noteworthy that the training set of Foggy Cityscapes \cite{sakaridis2018semantic} comprises 8925 data pairs, and Virtual KITTI \cite{gaidon2016virtual} includes 3452 data pairs, both exceeding the 1350 data pairs in SynFog. However, models trained on SynFog demonstrate superior generalization capabilities on real-world data compared to them, highlighting the crucial role of data quality for researchers.

\par \textbf{Comparison with SynFog-\(\beta\) synthesized by ASM.} With the ability to obtain pixel-accurate depth information for each image, we can eliminate the impact of inaccurate depth and focus solely on the fog simulation method itself. The foggy image is synthesized from its clear counterpart in SynFog according to Eq. \eqref{JT_model}. Specifically, we employ the same airlight estimation method used in the construction of Foggy Cityscapes \cite{sakaridis2018semantic}, which involves computing the median of all the 0.1\(\% \) pixels with the largest dark channel values, as proposed in \cite{tang2014investigating}. We use DehazeFormer \cite{song2023vision} to train on both SynFog and SynFog-\(\beta\), followed by testing the trained model on real-world foggy images. As shown in Fig. \ref{fig:4.4_ablation}, image output by model trained on SynFog exhibits a better visual appearance compared to that of SynFog-\(\beta\).

\subsection{Ablation Study of Sensor Noise}
\label{exp:ablation_study}
\par To demonstrate the importance of sensor noise simulation in our pipeline, we create SynFog-\(\alpha\) dataset. During the data generation process, the sensor-induced noise is set to zero, while all other settings remain consistent with those of the original SynFog dataset. We compare the defogging and detection performance of models trained on SynFog and SynFog-\(\alpha\) datasets, as shown in Fig. \ref{fig:ablation} and Tab. \ref{table:ablation1}-\ref{table:ablation2}. The model trained on SynFog that incorporates sensor noise produces more natural results with fewer artifacts and higher detection accuracy.

\begin{table}[t]
    \centering
    \small
    \setlength{\abovecaptionskip}{0.1cm}
    \caption{Impact of sensor noise on defogging performance.}
    \label{table:ablation1}
    \begin{tabular}{c|ccc}
        \bottomrule
         \multirow{2}{*}{Experimental setting} & \multicolumn{3}{c}{O-Haze \cite{ancuti2018haze}} \\
         \cline{2-4}
         & PSNR \(\uparrow\) & SSIM \(\uparrow\) & DHQI \(\uparrow\)\\
        \hline
        AECRNet+SynFog(w/o noise) & 15.16 & 0.5795 & 37.53\\
        AECRNet+SynFog(w/ noise) & \textbf{15.43} & \textbf{0.6116} & \textbf{44.46}\\
         \toprule
    \end{tabular}
    \vspace{-0.3cm}
\end{table}

\begin{table}[t]
    \centering
    \small
    \setlength{\abovecaptionskip}{0.1cm}
    \caption{Impact of sensor noise on detection performance. ``FZ" refers to Foggy Zurich. ``STF" represents Seeing Through Fog.}
    \label{table:ablation2}
    \begin{tabular}{c|c|c}
        \bottomrule
         \multirow{2}{*}{Experimental setting} & \multicolumn{1}{c|}{FZ \cite{dai2020curriculum}} & \multicolumn{1}{c}{STF \cite{bijelic2020seeing}}\\
         \cline{2-3}
         & mAP (\%) & mAP (\%)\\
        \hline
        AECRNet+SynFog(w/o noise) & 69.5 & 54.7\\
        AECRNet+SynFog(w/ noise) & \textbf{71.5} & \textbf{55.5}\\
         \toprule
    \end{tabular}
    \vspace{-0.6cm}
\end{table}

\begin{figure*}[t]
    \centering
	\includegraphics[width=0.98\textwidth]{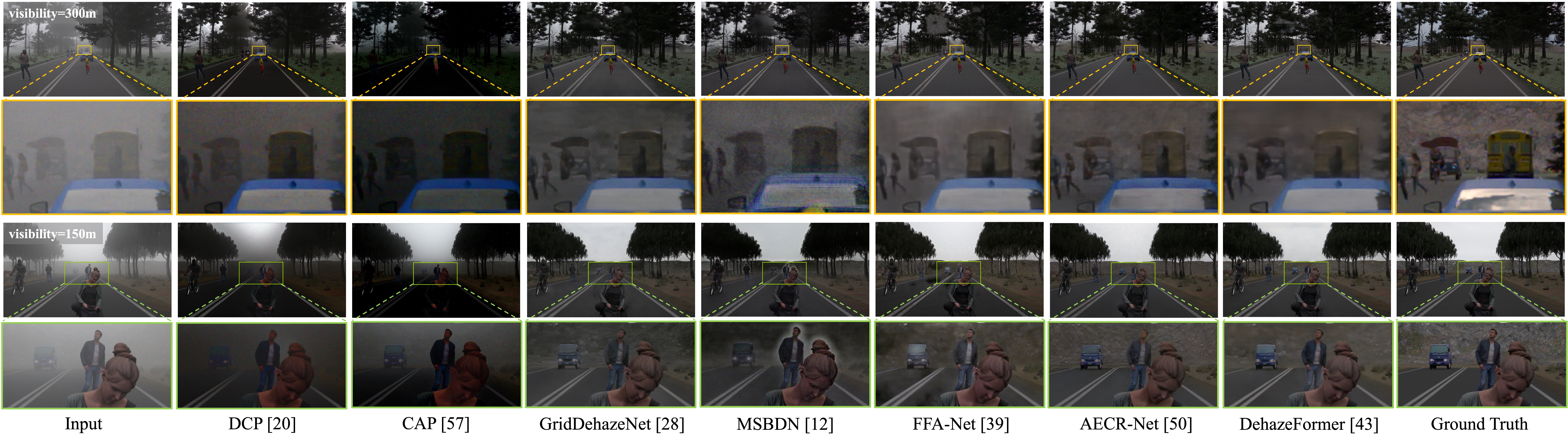}
	\caption{Qualitative comparisons of SOTA defogging methods on SynFog dataset with different fog levels. Zoom in for better view.}
	\label{fig:4.2_benchmark}
        \vspace{-0.5cm}
\end{figure*}

\vspace{-0.1cm}
\subsection{Algorithm Benchmarking}
\label{exp:benchmark}
\par We use SynFog test set to assess the aforementioned representative defogging methods, employing both full-reference and no-reference metrics to evaluate the defogged results. To illustrate the challenges presented by SynFog, we also provide evaluation results from the widely-used RESIDE dataset \cite{li2018benchmarking}. The full-scale quantitative comparison results are available in the supplementary material.

\par Generally, learning-based methods consistently outperform earlier algorithms based on natural or statistical priors in terms of PSNR, SSIM and LPIPS \cite{zhang2018unreasonable}. In particular, DehazeFormer \cite{song2023vision} attains the highest PSNR (27.96) and SSIM (0.923) values, while GridDehazeNet \cite{liu2019griddehazenet} achieves the best LPIPS (0.163) value on SynFog test set. Although the overall performance trend of the algorithms on SynFog is similar to that on RESIDE \cite{li2018benchmarking}, the evaluation scores are noticeably lower, highlighting the challenges posed by SynFog. Fig. \ref{fig:4.2_benchmark} depicts the qualitative comparison among these methods across different fog levels. Images restored by DCP \cite{he2010single} and CAP \cite{zhu2015fast} tend to appear excessively dark and noisy due to inaccurate airlight estimation and the lack of effective denoising capabilities. GridDehazeNet \cite{liu2019griddehazenet} lacks color recovery ability in dense fog, while the outputs of FFA-Net \cite{qin2020ffa} and DehazeFormer \cite{song2023vision} seem smooth. This comparison further indicates the substantial room for improvement in existing defogging methods, especially when applied to more realistic fog datasets.

\subsection{Discussion and Limitations}
\label{exp:discussion}
Experimental results reveal that models trained on SynFog, incorporating realistic physical characteristics of optics and sensors, exhibit a more robust performance on real-word fog datasets in terms of visual perception and detection accuracy. This outcome implies that integrating a physically-based end-to-end imaging process replicating real capture methods can significantly improve the generalization of trained models. However, given the time-consuming nature of the physically-based volumetric path tracing technique, the scale of SynFog is not as extensive as other synthetic fog datasets. Nonetheless, we are committed to expanding the scene contents in the future and exploring a broader range of scattering mediums beyond fog.

\vspace{-0.2cm}
\section{Conclusion}
\label{sec:conclusion}
\par In this paper, we present an end-to-end foggy image simulation pipeline. Our approach utilizes volumetric path tracing to model a more precise light scattering process with global illumination. By incorporating a physically-based camera processing pipeline that includes optics, sensor and image processing, we can closely mimic the authentic capture process under foggy conditions. Additionally, we develop a new synthetic fog dataset, SynFog, to facilitate the research on defogging. Comprehensive experiments have validated the authenticity and reliability of the SynFog dataset.

\vspace{0.2cm}

\par \noindent \textbf{Acknowledgements} This work is supported by 1) National Natural Science Foundation of China (Grant No. 62131011 and 62375233); 2) NSFC Young Scientists Fund (Grant No. 62302423). We thank Professor Chuxi Yang and Jiayue Xie for their help in preparing the scene creation. We also thank Professor Brian Wandell, Dr. Devesh Upadhyay, Dr. Alireza Rahimpour, and Devesh Shah for their valuable advice at the early stage of this project.
{
  \small
  \bibliographystyle{ieeenat_fullname}
  \bibliography{main}
}


\end{document}